\documentclass[5p,twocolumn]{elsarticle}

\usepackage{amssymb}
\usepackage{amsmath}
\usepackage{graphicx}
\usepackage{caption}
\usepackage{subcaption}
\usepackage{url}
\usepackage{booktabs}
\usepackage{multirow}
\usepackage{tabularx}
\usepackage{microtype}

\journal{Nuclear Physics B}

\begin{document}

\begin{frontmatter}

%% Title, authors and addresses

%% use the tnoteref command within \title for footnotes;
%% use the tnotetext command for theassociated footnote;
%% use the fnref command within \author or \affiliation for footnotes;
%% use the fntext command for theassociated footnote;
%% use the corref command within \author for corresponding author footnotes;
%% use the cortext command for theassociated footnote;
%% use the ead command for the email address,
%% and the form \ead[url] for the home page:
%% \title{Title\tnoteref{label1}}
%% \tnotetext[label1]{}
%% \author{Name\corref{cor1}\fnref{label2}}
%% \ead{email address}
%% \ead[url]{home page}
%% \fntext[label2]{}
%% \cortext[cor1]{}
%% \affiliation{organization={},
%%            addressline={}, 
%%            city={},
%%            postcode={}, 
%%            state={},
%%            country={}}
%% \fntext[label3]{}

\title{Knowledge Augmentation via Synthetic Data: A Framework for Real-World ECG Image Classification} %% Article title

%% use optional labels to link authors explicitly to addresses:
%% \author[label1,label2]{}
%% \affiliation[label1]{organization={},
%%             addressline={},
%%             city={},
%%             postcode={},
%%             state={},
%%             country={}}
%%
%% \affiliation[label2]{organization={},
%%             addressline={},
%%             city={},
%%             postcode={},
%%             state={},
%%             country={}}

\author[label1]{Xiaoyu Wang}
\author[label2]{Ramesh Nadarajah}
\author[label3]{Zhiqiang Zhang}
\author[label1]{David Wong}

%% Author affiliation
\affiliation[label1]{organization={Leeds Institute of Health Sciences},
            addressline={University of Leeds}, 
            city={Leeds},
            country={UK}}
\affiliation[label2]{organization={Leeds Institute of Cardiovascular and Metabolic Medicine},
            addressline={University of Leeds}, 
            city={Leeds},
            country={UK}}
\affiliation[label3]{organization={School of Electronic and Electrical Engineering},
            addressline={University of Leeds}, 
            city={Leeds},
            country={UK}}

%% Abstract
\begin{abstract}

In real-world clinical practice, electrocardiograms (ECGs) are often captured and shared as photographs. However, publicly available ECG data, and thus most related research, relies on digital signals. This has led to a disconnect in which computer assisted interpretation of ECG cannot easily be applied to ECG images. The emergence of high-fidelity synthetic data generators has introduced practical alternatives by producing realistic, photo-like, ECG images derived from the digital signal that could help narrow this divide. 

To address this, we propose a novel knowledge augmentation framework that uses synthetic data generated from multiple sources to provide generalisable and accurate interpretation of ECG photographs.
Our framework features two key contributions. 
First, we introduce a robust pre-processing pipeline designed to remove background artifacts and reduces visual differences between images. 
Second, we implement a two-stage training strategy: a Morphology Learning Stage, where the model captures broad morphological features from visually different, scan-like synthetic data, followed by a Task-Specific Adaptation Stage, where the model is fine-tuned on the photo-like target data.

We tested the model on the British Heart Foundation Challenge dataset, to classify five common ECG findings: myocardial infarction (MI), atrial fibrillation, hypertrophy, conduction disturbance, and ST/T changes. Our approach, built upon the ConvNeXt backbone, outperforms a single-source training baseline and achieved \textbf{1st} place in the challenge with an macro-AUROC of \textbf{0.9677}.
These results suggest that incorporating morphology learning from heterogeneous sources offers a more robust and generalizable paradigm than conventional single-source training.

\end{abstract}

% %%Graphical abstract
% \begin{graphicalabstract}
% %\includegraphics{grabs}
% \end{graphicalabstract}

% %%Research highlights
% \begin{highlights}
% \item We propose a synergistic framework to address the key gaps unique to real-world ECG photos (versus signals or scans). 
% It integrates a multi-stage pre-processing pipeline to narrow visual feature differences and a two-stage "Knowledge Augmentation" strategy to build robust domain priors.
% \item A core discovery is that ECG morphological knowledge is transferable across visual domains. 
% We demonstrate that pre-training on synthetic, visually mismatched 'scan-like' images significantly boosts diagnostic performance on 'photo-like' images. 
% This offers a new paradigm for addressing data scarcity and domain gaps in medical AI.
% \end{highlights}

%% Keywords

\end{frontmatter}

%% Add \usepackage{lineno} before \begin{document} and uncomment 
%% following line to enable line numbers
%% \linenumbers

%% main text
%%

%% Use \section commands to start a section
\section{Introduction}

Cardiovascular diseases (CVDs) remain the leading cause of mortality worldwide, accounting for an estimated 19.9 million deaths in 2024~\cite{20242024}. 
This escalating public health burden is expected to continue rising~\cite{BHF2024}. 
The 12-lead electrocardiogram (ECG) remains the gold-standard tool for frontline cardiac diagnostics~\cite{somani2021deep}. 
However, manual interpretation is time-consuming and heavily reliant on specialist expertise~\cite{winters2022time}. 
Automated ECG interpretation is therefore essential for improving diagnostic efficiency and patient outcomes.

Most existing AI systems for automated ECG analysis have concentrated on processing one-dimensional (1D) digital signals~\cite{liu2021deep} or clean, scanned images~\cite{abubaker2022detection}.  
In many clinical workflows, particularly in resource-limited settings or scenarios requiring rapid sharing, ECGs are frequently archived, shared, and reviewed not as digital signals, but as photographs of paper printouts~\cite{BHF2024}.
These photos introduce a distinct set of challenges not present in digital signals or scans, including perspective distortion, uneven illumination, shadows, physical wear (creases or stains), and background interference. 

Recent advances in high-fidelity synthetic data generation offer promising solutions to mitigate this gap. 
Tools such as GenECG~\cite{GenECG} can produce synthetic ECG images that emulate the visual characteristics of photo-like data with high realism. 
These resources have notably improved data availability for AI research; for example, the 2024 BHF Open Data Science Challenge dataset was derived entirely from GenECG~\cite{BHF2024}. 
However, relying on synthetic data brings an important but often overlooked risk.
When deep learning models are trained only on data from one generative source, they may learn the specific patterns or artifacts of that generator instead of the true, general features of ECG signals.
This leads to a Single-Source Limitation, where the model performs poorly when tested on data that comes from a different distribution than the one used by the synthetic engine.

In this paper, we propose and validate a Knowledge Augmentation (KA) deep learning framework designed to overcome this limitation and bridge the gap between synthetic training and real-world application. 
Rather than treating synthetic data as a monolithic source, our approach adopts a two-stage strategy to decouple general representation learning from target-specific optimisation.
Specifically, during the \textit{morphology learning stage}, the model acquires broad morphological knowledge from a visually distinct, synthetic dataset that contains clean images generated from multiple publicly-available ECG signal datasets, and subsequently adapts this knowledge to the photo-like target setting through \textit{task-specific adaptation stage}.
In addition, we introduce a pre-processing pipeline that aligns the visual characteristics between different image styles, making the cross-domain transfer more consistent and effective.

Our main contributions are:

\begin{itemize}
    \item A two-stage Knowledge Augmentation (KA) framework that uses clean, synthetic ECG images to improve the classification performance of real-world ECG photos. This framework shows clear gains over conventional single-source training.
    \item A robust pre-processing pipeline that reduces visual inconsistencies between ECG scans and photo data, enabling more stable and effective knowledge transfer.
    \item The proposed approach achieved first place in the 2024 BHF Open Data Science Challenge~\cite{BHF2024}, confirming its effectiveness on high-fidelity synthetic data and highlighting its potential for analysis of real-world ECG photographs.
\end{itemize}

The remainder of this paper is organised as follows.
Section~\ref{sec:related_work} reviews related work.
Section~\ref{sec:methodology} details our proposed methodology, followed by experimental results in Section~\ref{sec:results}.
Section~\ref{sec:discussion} discusses the findings and their implications, and finally, Section~\ref{sec:conclusion} concludes the paper.

\section{Related Work}
\label{sec:related_work}

This section summarises prior research in two main areas: the generation of synthetic ECG images and automated methods for ECG interpretation.

\subsection{Synthetic ECG Image Generation}
Large-scale, high-quality datasets of real ECG photographs are extremely limited. 
To address this challenge, several synthetic data generators have previously been developed. 
These can be broadly divided according to the visual characteristics of the images they produce.

\begin{figure}[htbp]
    \centering
    \captionsetup[subfigure]{justification=centering}
    \begin{subfigure}[b]{0.23\textwidth}
        \centering
        \includegraphics[width=\textwidth, height=2.5cm]{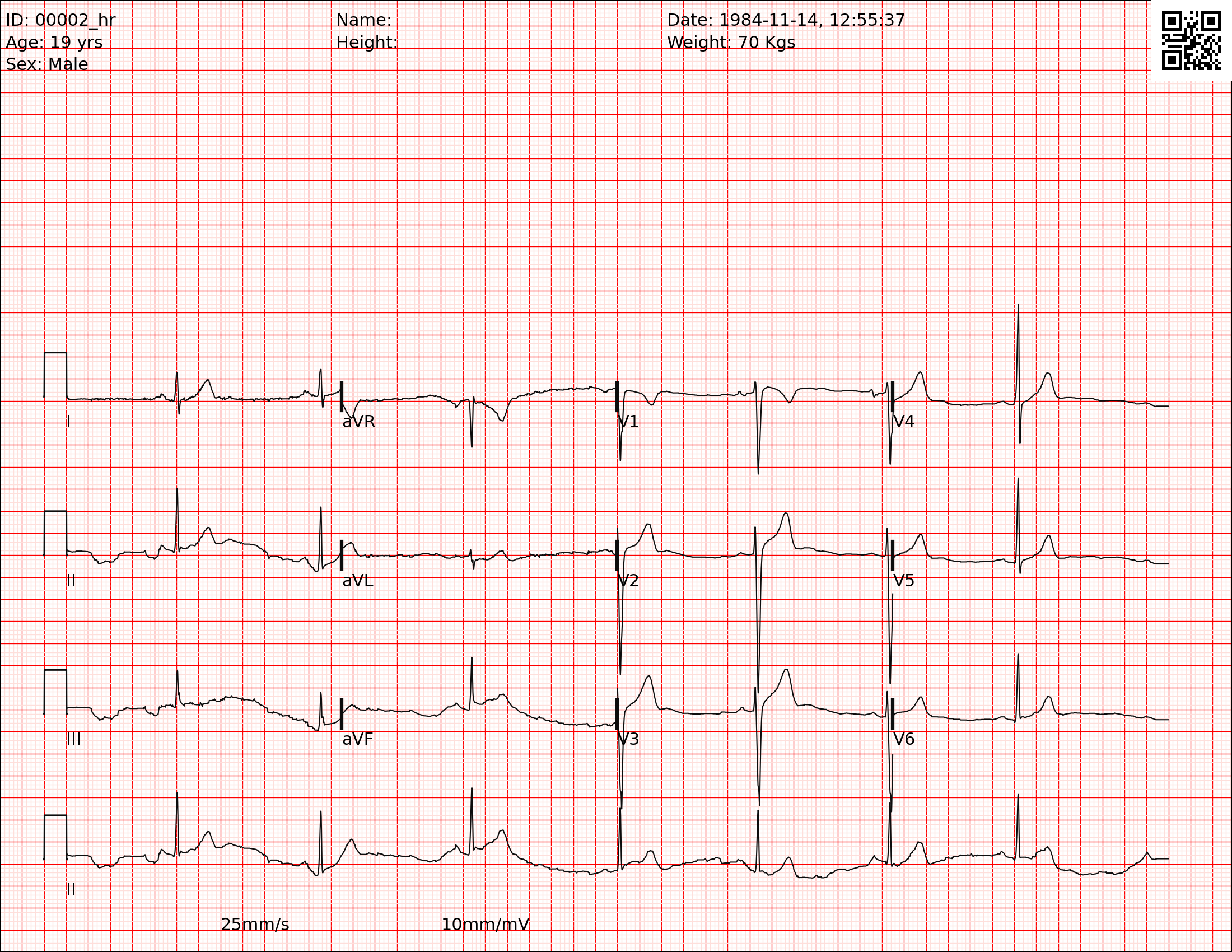}
        \caption{Images generated by \textbf{ECG-image-kit}~\cite{ecg-image-kit}}
        \label{fig:sub1}
    \end{subfigure}
    \hfill
    \begin{subfigure}[b]{0.23\textwidth}
        \centering
        \includegraphics[width=\textwidth, height=2.5cm]{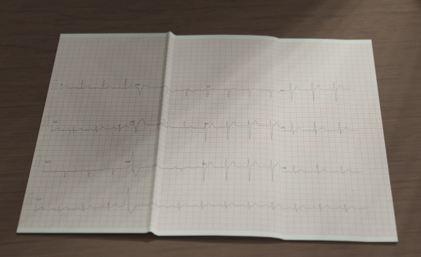}
        \caption{Images generated by \textbf{GenECG}~\cite{GenECG}}
        \label{fig:sub2}
    \end{subfigure}
    
    \caption{Examples of synthetic ECG images generated by different tools.}
    \label{fig:overall}
\end{figure}

Two representative tools illustrate current approaches to synthetic ECG image generation.
\textbf{ECG-image-kit}~\cite{ecg-image-kit} produces scan-like images by plotting ECG waveforms on paper-like backgrounds and adding two-dimensional artefacts such as handwritten text, wrinkles, and creases (Figure~\ref{fig:sub1}). 
The resulting images are planar and relatively clean, resembling scanned documents rather than photographs captured in clinical settings.

% In contrast, \textbf{GenECG}~\cite{GenECG} (Figure~\ref{fig:sub2}) generates photo-like images using 3D rendering techniques to simulate realistic photographic artefacts, including uneven lighting, shadows, perspective distortion, and cluttered backgrounds. 
In contrast, \textbf{GenECG}~\cite{GenECG} (Figure~\ref{fig:sub2}) adopts a physically-based rendering approach to bridge the domain gap between synthetic and real-world data. 
Unlike standard 2D synthesis, GenECG introduces realistic paper artefacts by employing texture synthesis algorithms to generate creases and wrinkles. 
Geometric distortions are modelled through affine and homography transformations to simulate varying camera angles, while environmental factors are addressed using gradient masks to create complex shadow and lighting effects. 
This multi-stage pipeline results in images that closely mimic the noise profile of clinical photographs.
The realism of these images has been confirmed through clinical expert evaluation, and GenECG was used to create the dataset for the 2024 BHF Open Data Science Challenge~\cite{BHF2024}.

\subsection{Automated Classification of ECG Images}
In automated ECG interpretation, several technical approaches have been explored. 
A large body of work focuses on classifying one-dimensional (1D) digital signals~\cite{somani2021deep, wasimuddin2020stages, baloglu2019classification, acharya2017application, zhao2020adaptive}. 
However, these methods are not directly compatible with clinical workflows where ECGs are often available only as images. 
Applying 1D-based models to photographs is therefore impractical, reinforcing the existence of a clear modality gap.

To address this issue, image-based methods have been proposed, which can be grouped into two main approaches:

\textbf{1. Signal Extraction Methods:}  
These methods attempt to reconstruct the 1D signal from the 2D image through a multi-stage pipeline, followed by classification using a conventional 1D model. 
Typical implementations employ deep learning architectures such as YOLOv7, U-Net, or ResUNet to segment the waveform, grid, and text components, after which the signal is algorithmically reconstructed~\cite{choudual, yoonsegmentation, summerton2024modular}.  
The main limitation of this approach is its fragility: such pipelines are designed primarily for clean, scan-like images. 
Their segmentation and reconstruction stages often fail when faced with the complex artefacts of photo-like images—such as shadows, folds, or perspective distortions. 
Moreover, errors from early stages propagate through the pipeline, substantially degrading final classification accuracy.

\begin{figure*}[t!]
    \centering
    \includegraphics[width=0.7\textwidth]{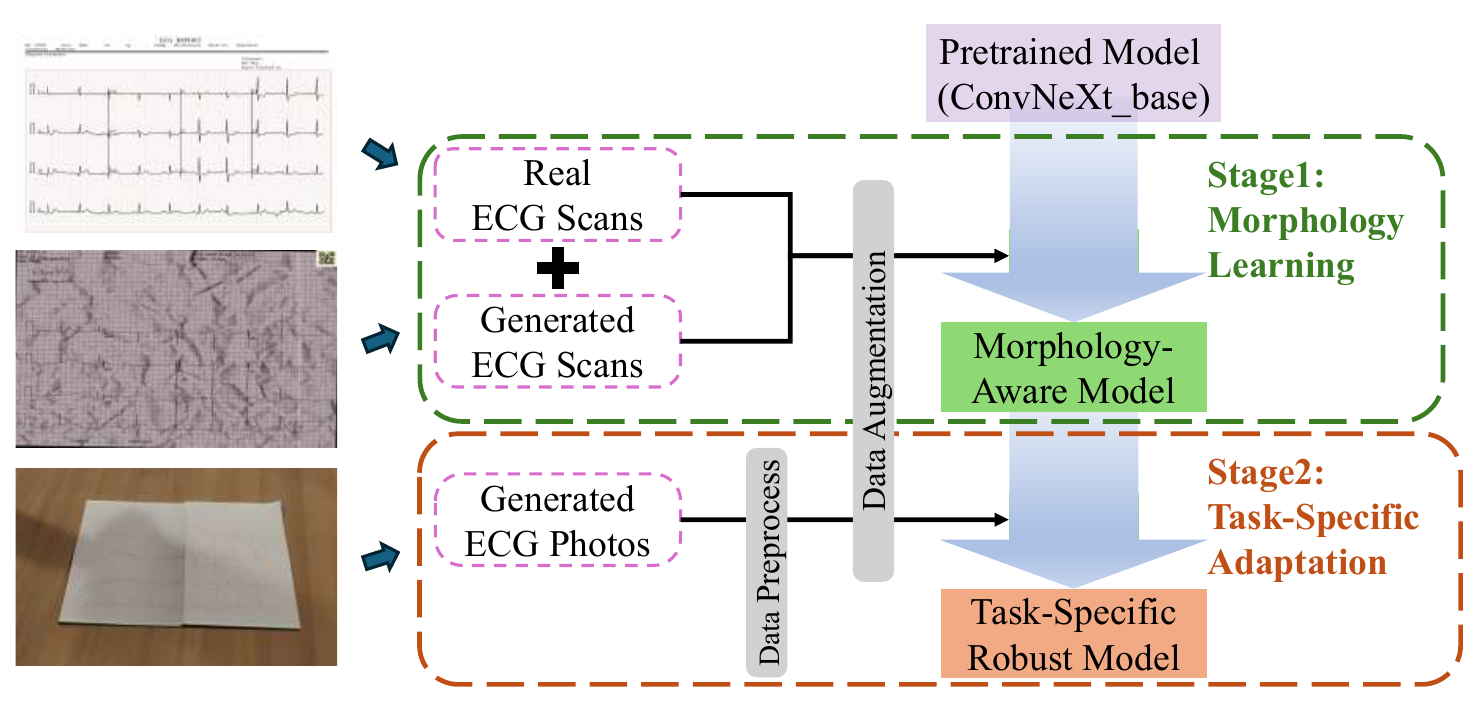}
    \vspace{-0.3cm}
    \caption{
    Overview of the proposed Knowledge Augmentation (KA) framework. 
    The process consists of a pre-processing pipeline followed by a two-stage training strategy. 
    Stage~I (Morphology Learning) trains on diverse ECG scans (including real ECG scans and generated ECG scans) to learn general ECG morphology. 
    Stage~II (Task-Specific Adaptation) fine-tunes the model on photo-realistic images (the BHF Challenge dataset - training data) for final classification.}
    \label{fig:overall-framework}
    \vspace{-0.3cm}
\end{figure*}

\textbf{2. Direct Classification Methods:}  
The second approach, adopted in this study, performs end-to-end classification by directly feeding the 2D image into a neural network (e.g., CNN or Transformer). 
Previous works have applied this strategy to tasks ranging from heartbeat-level classification~\cite{izci2019cardiac, jun2018ecg} to multi-label diagnosis using full 12-lead ECG images~\cite{nawaz2024covid, du2021fm, diasimage, antonidigital}.  
However, most of these studies focus on clean, scan-like images and have not been optimised for, or evaluated on, the more complex photo-like domain. 
As a result, these models tend to lack robustness when exposed to realistic artefacts such as uneven illumination, shadows, or geometric distortion, which are common in real-world photographs.

In summary, existing literature reveals a clear gap. 
1D signal methods are incompatible with image-based workflows, while signal extraction approaches are fragile due to their reliance on precise segmentation. 
Direct classification offers a more practical solution, but current methods still lack robustness under real-world photographic conditions. 
These limitations highlight the need for a framework capable of handling noisy, photo-like ECG images typically encountered in clinical practice.

\section{Methodology}
\label{sec:methodology}

To address the limitations of training on a single synthetic dataset, we propose a Knowledge Augmentation (KA) framework. 
The framework includes two components: (1) a two-stage training strategy that uses diverse synthetic data for general feature learning, and (2) a pre-processing pipeline that standardises image appearance. 
An overview is shown in Figure~\ref{fig:overall-framework}.

\subsection{Knowledge Augmentation Training Strategy}
The KA strategy splits training into two phases: learning general ECG features from datasets of ECG images (both real scans and generated scans) and adapting these features to photo-realistic data (the BHF Challenge dataset~\cite{BHF2024}).

\subsubsection{Stage I: Morphology Learning}
Stage~I trains the model on a set of visually diverse ECG images (comprising both real scans and generated scans) to learn general morphological patterns (e.g., P-waves, QRS complexes), independent of the visual style of any particular generator.

We constructed a training set using three scanned ECG datasets: ECG Images by Pondy et al.~\cite{pondy2023ecgimages}, COVID-19 Detection Using ECG~\cite{ibraa2023covid}, and the National Heart Foundation 2023 ECG Dataset~\cite{drkhaledmohsin2023ecg}. 
To further increase variation, we generated additional paper-style ECGs from the CODE15\% waveform dataset~\cite{CODE15} using the ecg-image-kit tool~\cite{ecg-image-kit}.

The ConvNeXt-Base model was trained for 50~epochs with a batch size of 32 using the AdamW optimiser~\cite{loshchilov2017decoupled} (learning rate $1\times10^{-4}$, weight decay $1\times10^{-2}$). 
We applied early stopping with a patience of 5 and used a cosine-annealing learning rate schedule:
\begin{equation}
\lambda(\text{step}) = 0.5 \left( 1 + \cos \left( 2 \pi \times \text{num\_cycles} \times \text{progress} \right) \right),
\end{equation}
where \emph{progress} denotes the training progress proportion. 

Class imbalance was addressed using weighted binary cross-entropy, with weights computed as the inverse of class prevalence. 
Data augmentation was implemented to all Stage I training images using Albumentations~\cite{info11020125}, including rotations, elastic deformation, perspective changes, brightness/contrast adjustments, and shadow simulation using Catmull–Rom splines~\cite{twigg2003catmull}.

\subsubsection{Stage II: Task-Specific Adaptation}
Stage~II fine-tunes the model on the photo-realistic data from the BHF Challenge dataset using five-fold cross-validation. 
Each fold was trained for 30~epochs with a learning rate of $1\times10^{-4}$ and batch size 32. 
The optimiser, loss, and early stopping settings were the same as in Stage~I. 
Final predictions on the hidden test set were produced by hard voting across the five models.

\begin{figure*}[t!]
    \centering
    \includegraphics[width=0.8\textwidth]{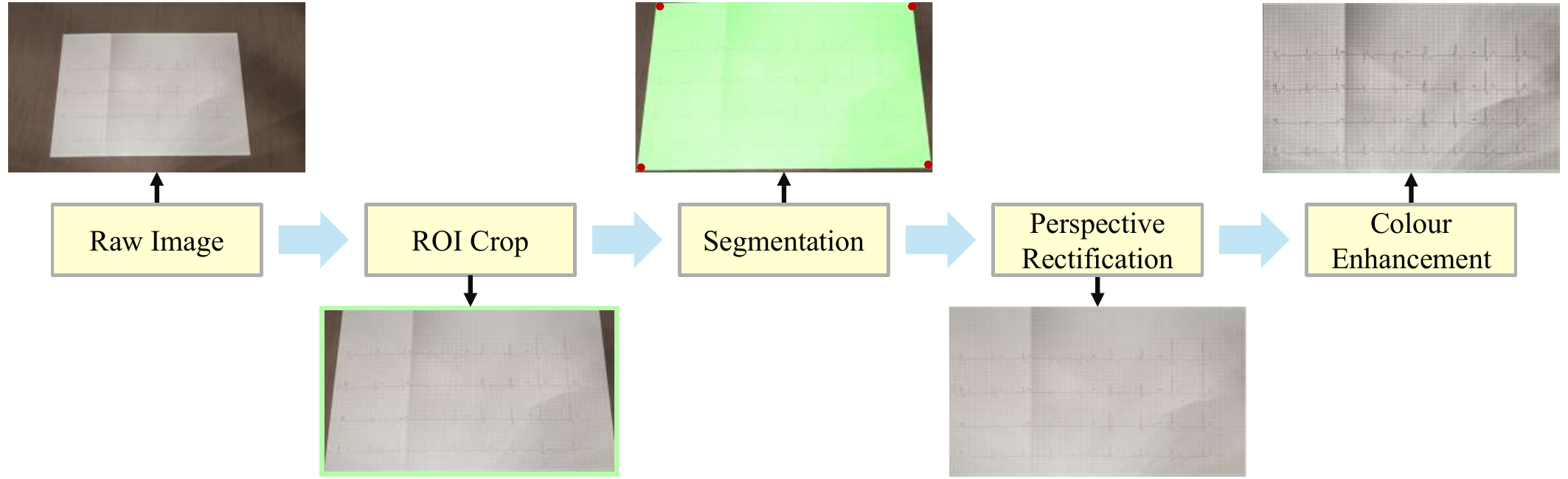}
    \vspace{-0.1cm}
    \caption{
        Overview of the data pre-processing workflow steps. 
        Step~1: Raw image input. 
        Step~2: ROI extraction using YOLO to remove surrounding background. 
        Step~3: Segmentation using SAM to isolate the paper sheet. 
        Step~4: Perspective rectification to correct the viewing angle. 
        Step~5: CLAHE colour enhancement. 
        Three strategies are evaluated based on these steps: (1) \textbf{Raw Input} (Step~1 only); (2) \textbf{ROI Crop + CLAHE} (Step~2 $\rightarrow$ Step~5); and (3) \textbf{Full Pipeline (Seg. + Rect.)} (Steps 1--5)..
        }
    \label{fig:preprocessing_pipeline}
    \vspace{-0.3cm}
\end{figure*}

\subsection{Data Pre-processing Pipeline}
The BHF Challenge images contain background clutter and perspective variation that differ from the synthetic datasets used in Stage~I. 
To address this visual gap, we evaluated a pre-processing pipeline designed to isolate the ECG signal and improve image consistency.

As illustrated in Figure~\ref{fig:preprocessing_pipeline}, the complete workflow consists of detection, segmentation, perspective transformation, and colour enhancement. 
Based on this workflow, we compared three distinct pre-processing strategies:
\begin{enumerate}
    \item \textbf{Raw Input:} The original images are used directly without any modification to establish a baseline.
    
    \item \textbf{ROI Crop + CLAHE:} This strategy utilises a fine-tuned YOLO detector~\cite{redmon2016you} to extract the Region of Interest (ROI), followed immediately by CLAHE~\cite{reza2004realization}. This approach removes background clutter while bypassing geometric transformations.
    
    \item \textbf{Full Pipeline (Seg. + Rect.):} This strategy incorporates all steps shown in Figure~\ref{fig:preprocessing_pipeline}. After cropping, the Segment Anything Model (SAM)~\cite{kirillov2023segment} isolates the ECG sheet, followed by a perspective rectification to a normalised top-down view before the final colour enhancement.
\end{enumerate}

\begin{figure}[t!]
    \centering
    \includegraphics[width=0.4\textwidth]{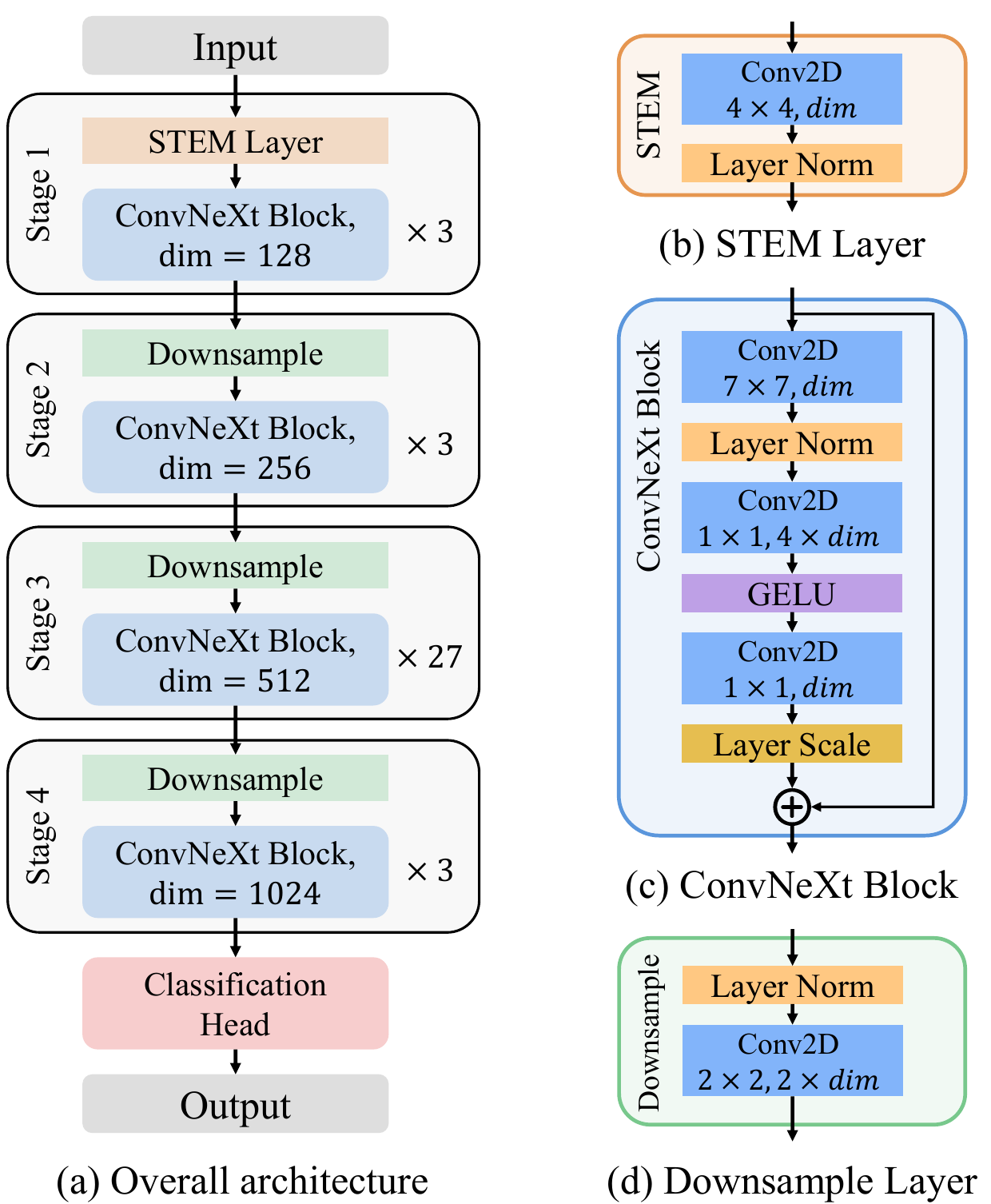}
    \caption{ConvNeXt architecture: (a) overall structure; (b) stem layer; (c) ConvNeXt block; (d) downsampling layer.}
    \label{fig:overall-architecture}
    \vspace{-0.5cm}
\end{figure}

\subsection{Model Architecture}

We adopt ConvNeXt-Base~\cite{liu2022convnet} as the backbone (Figure~\ref{fig:overall-architecture}). 
This architecture was selected based on its proven efficacy in processing medical imaging data, notably demonstrated by its success in previous ECG image classification tasks~\cite{dias2024image, reyna2024digitization}, which utilised a similar convolutional backbone.
To further validate this choice for our specific application, we conducted comparative experiments against other state-of-the-art architectures (ResNet-50, ViT, and Swin Transformer), identifying ConvNeXt as the optimal model (detailed results are provided in Section~\ref{sec:results}).

The model includes a patch-based stem, a hierarchical 1:1:9:1 depth configuration, inverted bottleneck blocks~\cite{sandler2018mobilenetv2}, and large $7\times7$ kernels to expand the receptive field. 
We use the \texttt{timm}~\cite{rw2019timm} implementation pre-trained on ImageNet-22k~\cite{deng2009imagenet}. 
Architecture parameters are channel dimensions $C=(128,256,512,1024)$ and block counts $B=(3,3,27,3)$.

\subsection{Dataset and Experimental Setup}

\subsubsection{Task Definition and Evaluation Metrics}

\begin{table*}[t!]
    \centering
    \caption{Detailed breakdown of the datasets used in Stage I. Vertical and horizontal lines are added to clearly distinguish between data sources and pathology counts.}
    \label{tab:stage1_details}
    \begin{tabularx}{0.85\textwidth}{l|c|c|X|r}
        \hline
        \hline
        \textbf{Dataset} & \textbf{Type} & \textbf{Total Size} & \textbf{Pathology Category} & \textbf{Count} \\
        \hline
        \multirow{4}{*}{\textbf{ECG Images}~\cite{pondy2023ecgimages}} & \multirow{4}{*}{Real Scans} & \multirow{4}{*}{1,376} 
        & Myocardial Infarction (MI) & 351 \\
        & & & Abnormal Heartbeat & 345 \\
        & & & History of MI & 284 \\
        & & & Normal & 396 \\
        \hline
        
        \multirow{5}{*}{\textbf{COVID-19 ECG}~\cite{ibraa2023covid}} & \multirow{5}{*}{Real Scans} & \multirow{5}{*}{1,932} 
        & Myocardial Infarction (MI) & 74 \\
        & & & Abnormal Heartbeat & 546 \\
        & & & History of MI & 203 \\
        & & & Normal & 859 \\
        & & & COVID-19 Positive & 250 \\
        \hline
        
        \multirow{4}{*}{\textbf{NHF 2023}~\cite{drkhaledmohsin2023ecg}} & \multirow{4}{*}{Real Scans} & \multirow{4}{*}{2,898} 
        & Myocardial Infarction (MI) & 716 \\
        & & & Abnormal Heartbeat & 814 \\
        & & & History of MI & 516 \\
        & & & Normal & 852 \\
        \hline
        
        \multirow{4}{*}{\textbf{CODE15\%}~\cite{CODE15}} & \multirow{4}{*}{Generated} & \multirow{4}{*}{345,779} 
        & 1st Degree AV Block & 5,716 \\
        & & & Conduction Defects & 15,698 \\
        & & & Sinus Brady/Tachycardia & 13,189 \\
        & & & Atrial Fibrillation (AF) & 7,033 \\
        \hline
        \hline
    \end{tabularx}
    \vspace{-0.4cm}
\end{table*}

The primary objective of this study was to classify five non-mutually exclusive ECG interpretations: myocardial infarction (MI), atrial fibrillation (AF), hypertrophy (HYP), conduction disturbance (CD), and ST/T changes (STTC). 
Evaluation was performed on a hidden, held-out portion of the BHF challenge dataset.
As this is a multi-label classification problem, the primary metric used to assess performance was the macro-averaged Area Under the Receiver Operating Characteristic curve (macro-AUROC), calculated as the arithmetic mean of the individual class AUROCs.

\subsubsection{Dataset}

To validate the proposed Knowledge Augmentation framework, we utilised a comprehensive set of data sources, categorised into source datasets for morphology learning and a target dataset for task-specific adaptation.

\noindent\textbf{Morphology Learning Datasets (Stage I):}
As detailed in Section 3.1.1, the training stage I utilised a large-scale composite dataset ($N \approx 352,000$) comprising real scans~\cite{pondy2023ecgimages, ibraa2023covid, drkhaledmohsin2023ecg} and synthetic paper-style images~\cite{CODE15}. 
A detailed breakdown of these datasets, including their types, sample sizes, and pathology distributions, is presented in Table~\ref{tab:stage1_details}.
This aggregation ensures the model is exposed to a diverse range of signal morphologies and visual artefacts before the subsequent task-specific adaptation in Stage II.

\noindent\textbf{Task-Specific Dataset (Stage II):}
We use the dataset from the \textbf{BHF Data Science Centre ECG Challenge}~\cite{bhf-data-science-centre-ecg-challenge}, 
which contains 21,799 12-lead ECG images generated from PTB-XL~\cite{PTB-XL} using GenECG~\cite{GenECG}. 
The dataset is split into 15,009 training, 3,219 validation, and 3,219 test samples. 
The class distribution for this dataset is shown in Table~\ref{tab:sample_distribution}.

\begin{table}[htbp]
    \centering
    \caption{Number and proportion of positive samples per diagnosis in the training set.}
    \label{tab:sample_distribution}
    \begin{tabular}{lc}
        \hline
        \hline
        \textbf{Diagnosis} & \textbf{No.(\%)}\\
        \hline
        MI   & 3,819 (25.44\%)\\
        AF   & 1,033 (6.88\%)\\
        HYP  & 1,850 (12.33\%)\\
        CD   & 3,137 (20.90\%)\\
        STTC & 3,640 (24.25\%)\\
        \hline
        \hline
    \end{tabular}
    \vspace{-0.5cm}
\end{table}

% \textcolor{red} In our primary expriment, we assessed the ability of the trained model to classify five non-mutually exclusive ECG interpretations: myocardial infarction (MI), atrial fibrillation (AF), hypertrophy (HYP), conduction disturbance (CD), and ST/T changes (STTC). We used the a hidden held-out portion of the BHF challenge dataset. The metric used to assess performance was the macro-AUC, that is the arithmetic mean of the individual class AUCs.

\subsubsection{Ablation Studies}

To comprehensively evaluate the proposed framework, three ablation studies were conducted:

\textbf{Ablation Study I: Backbone Model Selection.} 
To identify the most effective feature extractor, we conducted a comparative analysis of four backbone architectures: the standard CNN baseline (ResNet-50), Transformer-based models (ViT, Swin Transformer), and the modern CNN variant (ConvNeXt).

\textbf{Ablation Study II: Knowledge Augmentation.} 
We assessed the efficacy of the full two-stage strategy (Morphology Learning + Task-Specific Adaptation) by comparing it against a baseline model trained exclusively on the BHF dataset.

\textbf{Ablation Study III: Pre-processing Strategy.} 
We investigated how data pre-processing shapes model behaviour by evaluating three conditions: Raw Input, ROI Crop + CLAHE, and the Full Pipeline (Seg. + Rect.). 
Specifically, we analysed the impact of background noise and geometric transformation on performance (e.g., AUROC).

% Two ablation studies were conducted:

% \textbf{1. Knowledge Augmentation:}  
% Comparing the full two-stage strategy (Morphology Learning + Task-Specific Adaptation) with a baseline trained only on the BHF dataset.

% \textbf{2. Pre-processing:}  
% Evaluating the three pre-processing methods (None, \textbf{Crop \& Enhance}, Segment \& Transform).

\subsubsection{Implementation Details}

Experiments were implemented in PyTorch~\cite{imambi2021pytorch} using two NVIDIA T4 GPUs. 
Unless stated otherwise, hyperparameters were kept fixed across experiments: 
learning rate $1\times10^{-4}$ with a 0.5 decay factor every 7~epochs, 
30~epochs total, dropout 0.2, and batch size 32. 
Performance is reported using AUROC. 
Code is available at \url{https://github.com/Nicholas0917/BHF-Challenge-2024.git}.

\goodbreak

\section{Results}
\label{sec:results}

This section presents the experimental results on the BHF Challenge dataset used to validate the proposed Knowledge Augmentation (KA) framework. 
We first report the \textbf{overall performance} of the complete system, followed by a series of \textbf{ablation studies} that assess the contribution of three key components: the backbone architecture, the KA training strategy, and the preprocessing pipeline. 
All results are evaluated using the AUROC metric on both the \textit{Public Validation Set} (used during the challenge) and the \textit{Hidden Test Set} (used for final ranking).

\subsection{Performance of the Proposed Framework}
\label{sec:main_results}

We first evaluate the integrated framework, which combines a \textbf{ConvNeXt} backbone, a \textbf{Two-stage (Real + Synthetic)} training procedure, and an ROI Crop + CLAHE pre-processing pipeline.
Table~\ref{tab:proposed_performance} summarises the performance using two synthetic data sources: CODE15\% and PTB-XL.

% We begin by evaluating the integrated framework combining the \textbf{ConvNeXt} backbone, the \textbf{Two-stage (Real + Synthetic)} training procedure, and the  ROI Crop + CLAHE pre-processing pipeline. 
% Table~\ref{tab:proposed_performance} summarises the performance obtained using two synthetic data sources (CODE15\% and PTB-XL). 
% Among them, the variant trained with PTB-XL synthetic data achieves the highest performance on both evaluation sets, demonstrating the robustness and effectiveness of our full pipeline.

\begin{table}[htbp]
\centering
\caption{Performance of the proposed framework after hyperparameter optimisation. "Public" denotes the public validation set used during the competition, whereas "Hidden" denotes the hidden test set used for final evaluation.}
\label{tab:proposed_performance}
\begin{tabular}{lcc}
\hline
\hline
\textbf{Synthetic Source} & \textbf{Public} & \textbf{Hidden} \\
\hline
CODE15\%   & 0.9472 & 0.9430 \\
PTB-XL     & \textbf{0.9688} & \textbf{0.9677} \\
\hline
\hline
\end{tabular}
\vspace{-0.2cm}
\end{table}

To further validate the competitiveness of our proposed method, we compare our best-performing model (trained with PTB-XL) against the top entries on the official competition leaderboard. 
As shown in Table~\ref{tab:leaderboard}, our framework outperforms the competing solutions, securing the \textbf{1st place} on the decisive Hidden test set.

\begin{table}[htbp]
\centering
\caption{Comparison with the top 5 teams on the competition leaderboard~\cite{kaggle_leaderboard}. Our proposed framework achieves the highest scores on both Public and Hidden sets.}
\label{tab:leaderboard}
\begin{tabular}{lcc}
\hline
\hline
\textbf{Team} & \textbf{Public} & \textbf{Hidden} \\
\hline
\textbf{XiaoyuWang\_12 (Ours)} & \textbf{0.9688} & \textbf{0.9677} \\
DERI\_QMUL & 0.9539 & 0.9490 \\
UniOfLiverpool & 0.9509 & 0.9475 \\
PIMI~\cite{buyuksolak2025pic2diagnosis} & 0.9495 & 0.9421 \\
Michael Ibrahim & 0.9432 & 0.9382 \\
\hline
\hline
\end{tabular}
\vspace{-0.2cm}
\end{table}

\subsection{Ablation Study I: Backbone Model Selection}

To justify the choice of backbone architecture, we compared four representative deep learning models under identical training conditions. 
Table~\ref{tab:model_performance} reports the AUROC scores across both evaluation sets. 
The results show that \textbf{ConvNeXt} consistently achieves the best performance, outperforming ResNet-50, ViT, and Swin Transformer. 
Accordingly, ConvNeXt was selected as the default backbone for all subsequent experiments.

\begin{table}[htbp]
\centering
\caption{Performance comparison of different backbone models on the BHF Challenge dataset.}
\label{tab:model_performance}
\begin{tabular}{lcc}
\hline
\hline
\textbf{Model} & \textbf{Public} & \textbf{Hidden} \\
\hline
ResNet-50        & 0.7159 & 0.6997 \\
ViT              & 0.8197 & 0.8209 \\
Swin             & 0.8722 & 0.8675 \\
ConvNeXt         & \textbf{0.8750} & \textbf{0.8683} \\
\hline
\hline
\end{tabular}
\vspace{-0.5cm}
\end{table}

\subsection{Ablation Study II: Empirical Validation of the Knowledge Augmentation Framework}

To evaluate the contribution of the KA framework, we conducted an ablation study designed to isolate the effect of the two-stage training strategy from that of synthetic data augmentation. 
Three training configurations were examined:
\begin{enumerate}
    \item \textbf{Single-stage baseline:} Task-specific training only.
    \item \textbf{KA framework (Real data only):} The two-stage pipeline without synthetic data.
    \item \textbf{KA framework (Real + Synthetic):} The full KA configuration combining real and synthetic ECG data.
\end{enumerate}

As shown in Table~\ref{tab:ablation_ka}, the two-stage strategy using real data outperforms the single-stage baseline, and adding synthetic data yields additional improvement on the Public Validation Set.

\begin{table}[t!]
    \centering
    \caption{Ablation study on the Knowledge Augmentation (KA) strategy. `Real' and `Synth.' denote the data sources used in Stage I.}
    \label{tab:ablation_ka}
    \begin{tabularx}{0.95\columnwidth}{X c c c c} 
        \hline
        \hline
        & \multicolumn{2}{c}{\textbf{Stage I Data}} & \multicolumn{2}{c}{\textbf{AUROC}} \\
        \cmidrule(lr){2-3} \cmidrule(lr){4-5}
        \textbf{Method} & \textbf{Real} & \textbf{Synth.} & \textbf{Public} & \textbf{Hidden} \\
        \hline
        Baseline & - & - & 0.8339 & 0.8172 \\
        KA (Real)     & \checkmark & - & 0.8677 & \textbf{0.8729} \\
        \textbf{KA (Full)}   & \checkmark & \checkmark & \textbf{0.8750} & 0.8683 \\
        \hline
        \hline
    \end{tabularx}
    \vspace{-0.3cm}
\end{table}

\subsection{Ablation Study III: Empirical Validation of the Preprocessing Enabler}

Finally, to assess the impact of preprocessing, we evaluated the three strategies described in Section~3.2: (1) \textbf{Raw Input}, (2) \textbf{ROI Crop + CLAHE}, and (3) \textbf{Full Pipeline (Seg. + Rect.)}. 
The AUROC results, shown in Table~\ref{tab:ablation_preprocess}, indicate that the \textbf{ROI Crop + CLAHE} strategy achieves the strongest performance across both evaluation sets, outperforming both the raw input and the more complex \textbf{Full Pipeline (Seg. + Rect.)}. 
Based on these findings, \textbf{ROI Crop + CLAHE} was adopted as the default preprocessing strategy for the full framework.

% Finally, to assess the impact of preprocessing, we evaluated the three strategies described in Section~3.2: (1) \textbf{None}, (2) \textbf{Crop \& Enhance}, and (3) \textbf{Full Pipeline (Segment \& Transform)}. 
% The AUROC results, shown in Table~\ref{tab:ablation_preprocess}, indicate that the \textbf{Crop \& Enhance} strategy achieves the strongest performance across both evaluation sets, outperforming both the raw input and the more complex transformation method. 
% Based on these findings, \textbf{Crop \& Enhance} was adopted as the default preprocessing strategy for the full framework.

\begin{table}[htbp]
    \centering
    \caption{Quantitative evaluation of preprocessing strategies. `Seg. + Rect.' denotes the segmentation and perspective rectification steps included in the full pipeline.}
    \label{tab:ablation_preprocess}
    \setlength{\tabcolsep}{4pt}
    \begin{tabularx}{0.9\columnwidth}{X c c}
        \hline
        \hline
        \textbf{Method} & \textbf{Public} & \textbf{Hidden} \\
        \hline
        Raw Input & 0.7703 & 0.7531 \\
        \addlinespace[0.5ex]
        \textbf{ROI Crop + CLAHE} & \textbf{0.8750} & \textbf{0.8683} \\
        \addlinespace[0.5ex]
        Full Pipeline (Seg. + Rect.) & 0.8359 & 0.8339 \\
        \hline
        \hline
    \end{tabularx}
    \vspace{-0.3cm}
\end{table}

\section{Discussion}
\label{sec:discussion}

In the \textit{Introduction}, we identified the limitations of using single-source synthetic data and noted the challenges of handling heterogeneous ECG images. 
The main results in Section~\ref{sec:main_results} show that our proposed solution—the Knowledge Augmentation (KA) framework combined with a robust pre-processing pipeline—effectively addresses these issues, achieving an AUROC of 0.9677. 
In this section, we examine the mechanisms behind this performance.

\subsection{Performance of the Proposed Framework and Limitations}

Our final model, optimised through hyperparameter tuning (Table~\ref{tab:ablation_ka}), achieved an AUROC of 0.9677 when PTB-XL was used as the source dataset during the KA pre-training stage.

A key concern regarding this result is the possibility of \textbf{data leakage}, as the target GenECG dataset is statistically derived from PTB-XL. 
To ensure that the model was not simply memorising source-specific patterns, we performed a "stress test" using CODE15\% as the source dataset (Table~\ref{tab:proposed_performance}).

CODE15\% has no overlap with PTB-XL, yet the model still obtained an AUROC of 0.9430. 
This strong performance on an independent dataset rules out data leakage as the main explanation. 
Instead, it provides clear evidence of the \textbf{generalisability} of the KA framework. 
The findings indicate that the first training stage enables the model to learn universal morphological representations that transfer well across different data sources, rather than overfitting to the characteristics of a single dataset.

\subsection{Analysis of the Knowledge Augmentation Framework}
The results from Ablation Study I (Table~\ref{tab:ablation_ka}) provide clear evidence for the value of the KA component. 
The KA method (AUROC 0.8750) significantly outperforms the method without the \textbf{morphology learning stage} (AUROC 0.8339). 

This performance gap demonstrates the effectiveness of the proposed two-stage training strategy. It confirms that the first stage of training, Morphology Learning on a diverse set of synthetic ECG data, is beneficial. This process helps the model to learn \textbf{robust and relevant morphological features}, which improves performance on the final classification task.

\subsection{Analysis of Pre-processing Strategies}

Ablation Study III (Table~\ref{tab:ablation_preprocess}) shows that choosing a pre-processing strategy involves balancing geometric alignment with data integrity. 
The Raw Input condition (AUROC 0.7703) performs poorly due to background noise, while comparing ROI Crop + CLAHE (AUROC 0.8750) and the Full Pipeline (Seg. + Rect.) (AUROC 0.8359) provides further insight into how pre-processing shapes model behaviour.

We attribute the advantage of the ROI Crop + CLAHE strategy to two main factors: differences in feature attention and the balance between robustness and complexity.

\subsubsection{Divergence in Feature Attention: Global Context vs Local Morphology}

\begin{figure*}[t!]
    \centering
    \begin{subfigure}[b]{0.3\textwidth}
        \centering
        \includegraphics[width=\textwidth]{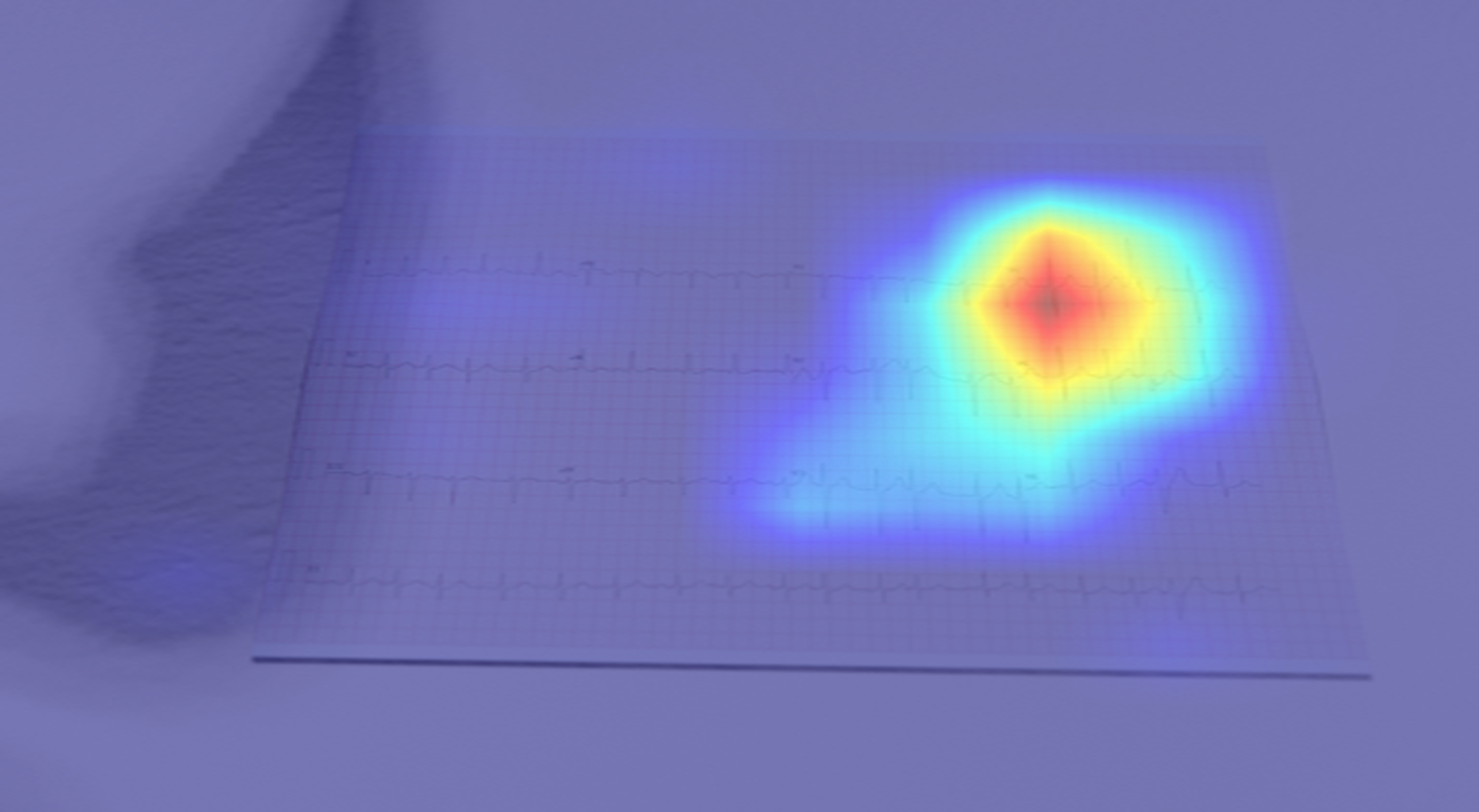}
        \caption{AF: Raw Input}
        \label{fig:all_af_none}
    \end{subfigure}
    \hfill
    \begin{subfigure}[b]{0.3\textwidth}
        \centering
        \includegraphics[width=\textwidth]{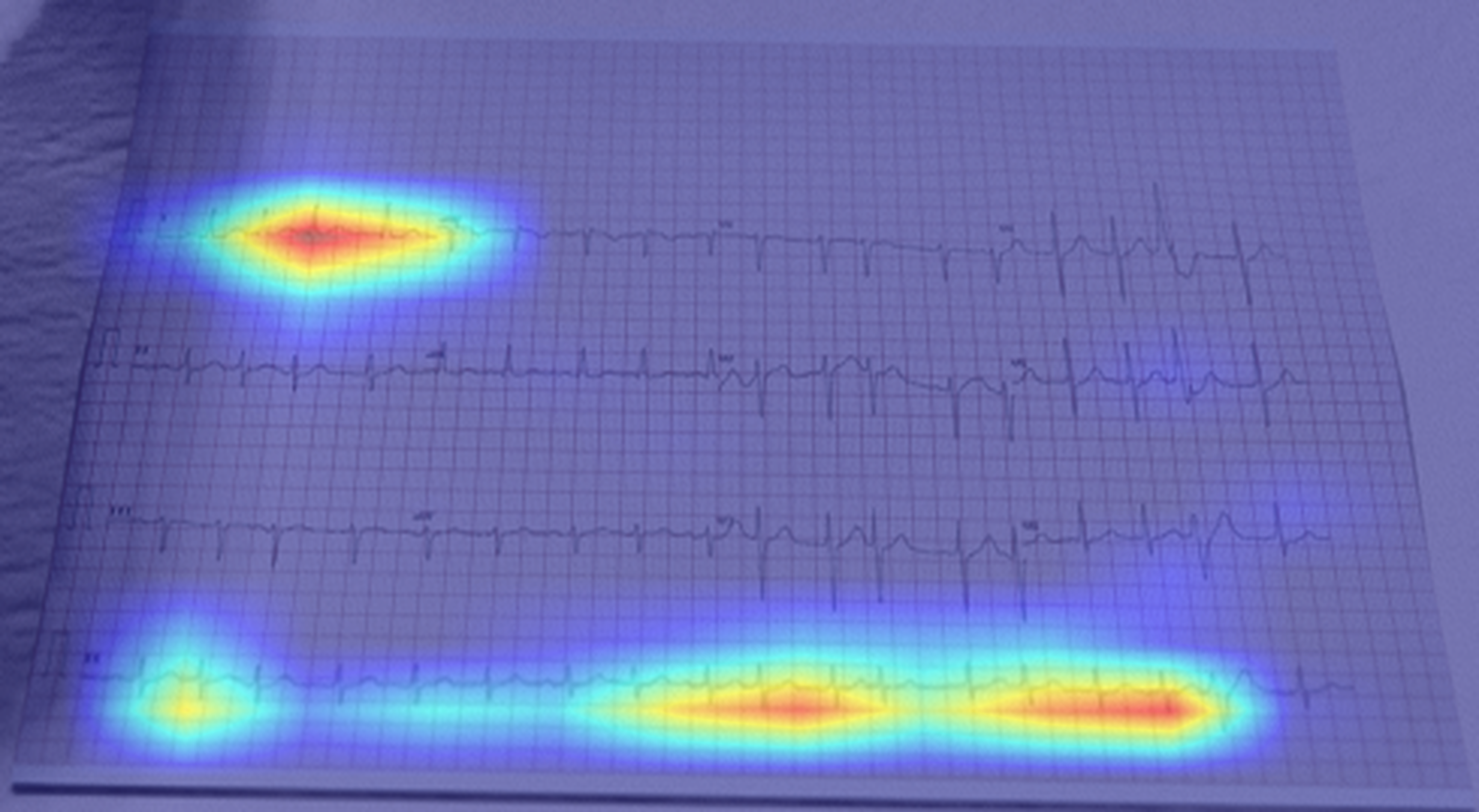}
        \caption{AF: ROI Crop + CLAHE}
        \label{fig:all_af_crop}
    \end{subfigure}
    \hfill
    \begin{subfigure}[b]{0.3\textwidth}
        \centering
        \includegraphics[width=\textwidth]{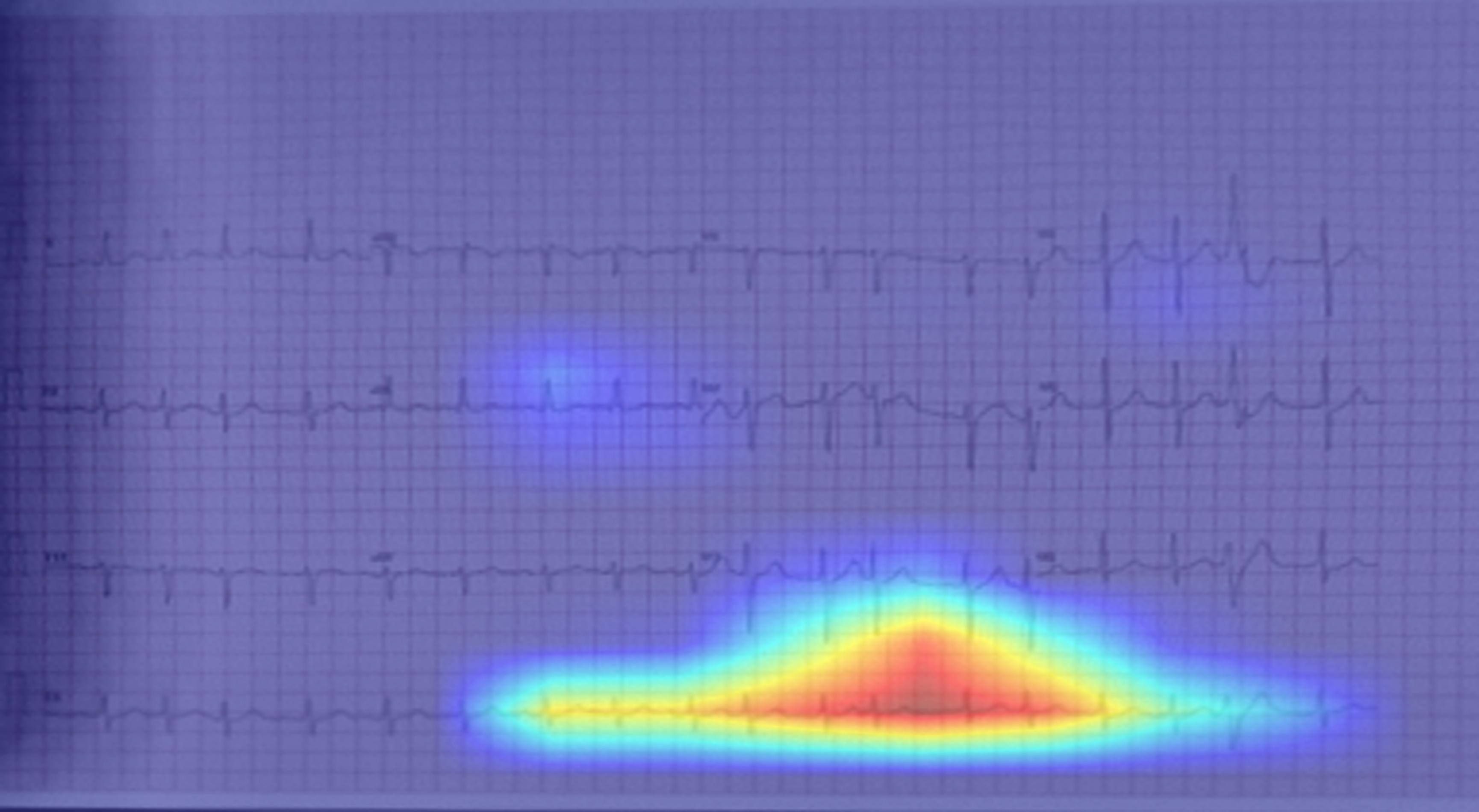}
        \caption{AF: Full Pipeline (Seg. + Rect.)}
        \label{fig:all_af_full}
    \end{subfigure}
    
    \vspace{1em}
    
    \begin{subfigure}[b]{0.3\textwidth}
        \centering
        \includegraphics[width=\textwidth]{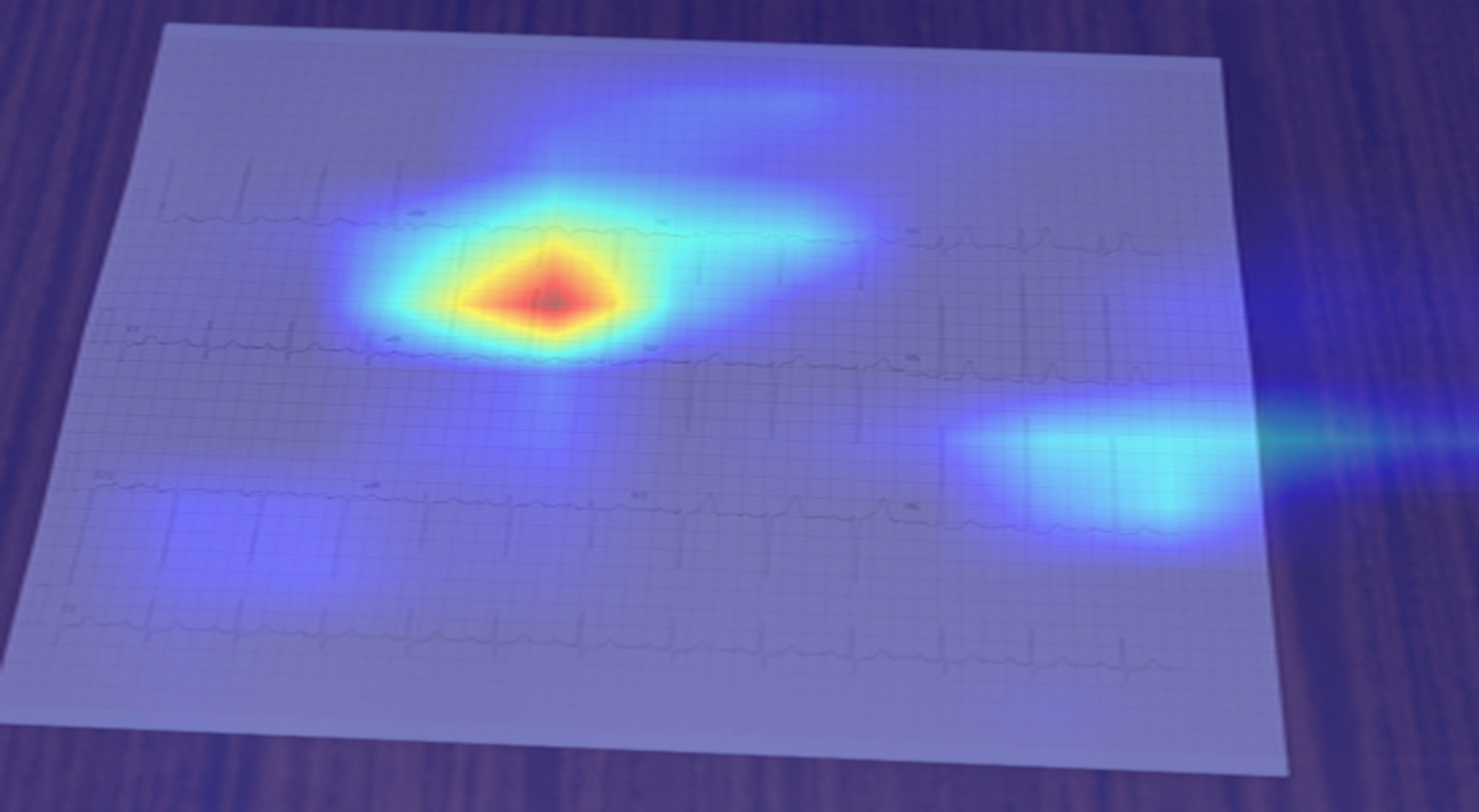}
        \caption{HYP: Raw Input}
        \label{fig:all_hyp_none}
    \end{subfigure}
    \hfill
    \begin{subfigure}[b]{0.3\textwidth}
        \centering
        \includegraphics[width=\textwidth]{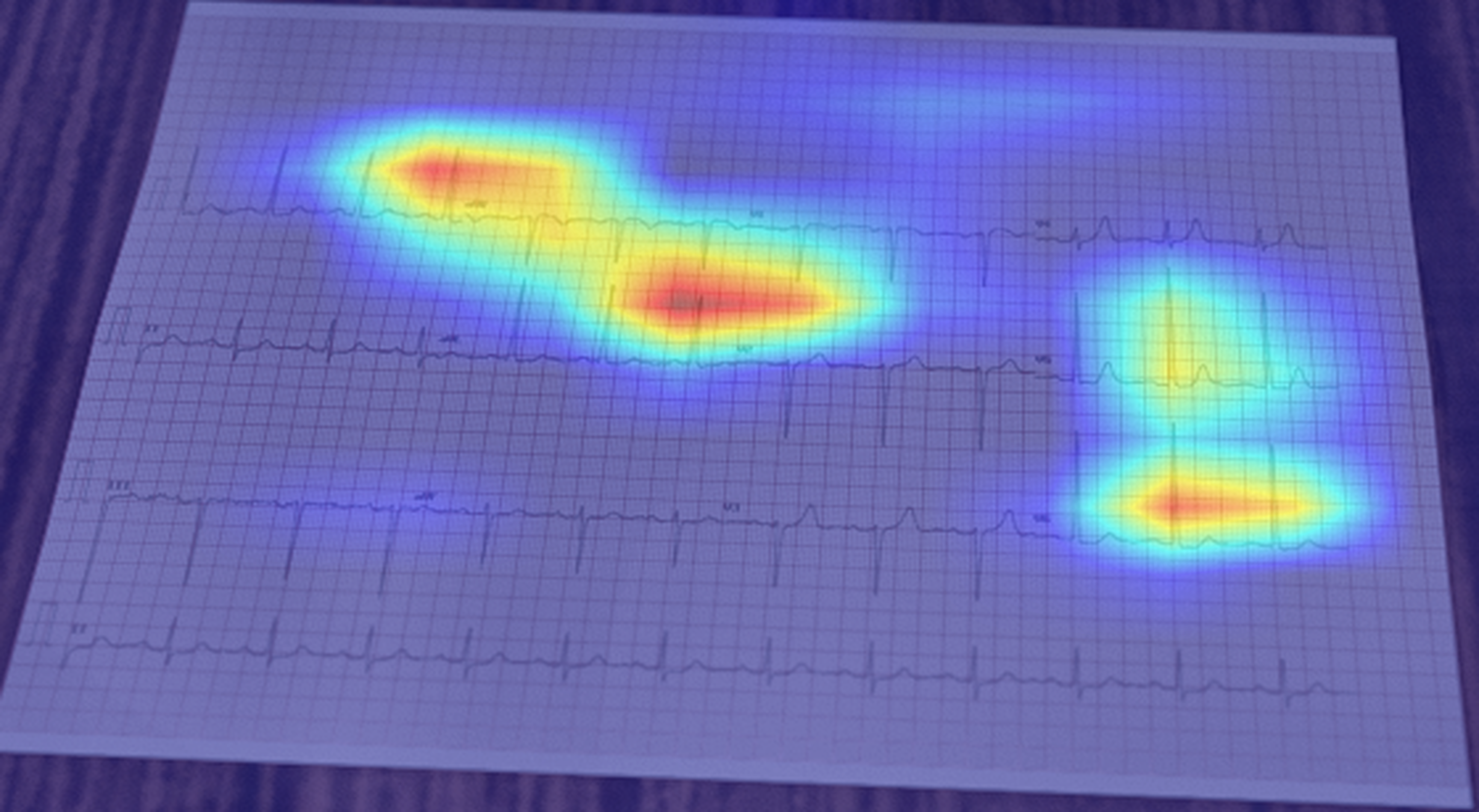}
        \caption{HYP: ROI Crop + CLAHE}
        \label{fig:all_hyp_crop}
    \end{subfigure}
    \hfill
    \begin{subfigure}[b]{0.3\textwidth}
        \centering
        \includegraphics[width=\textwidth]{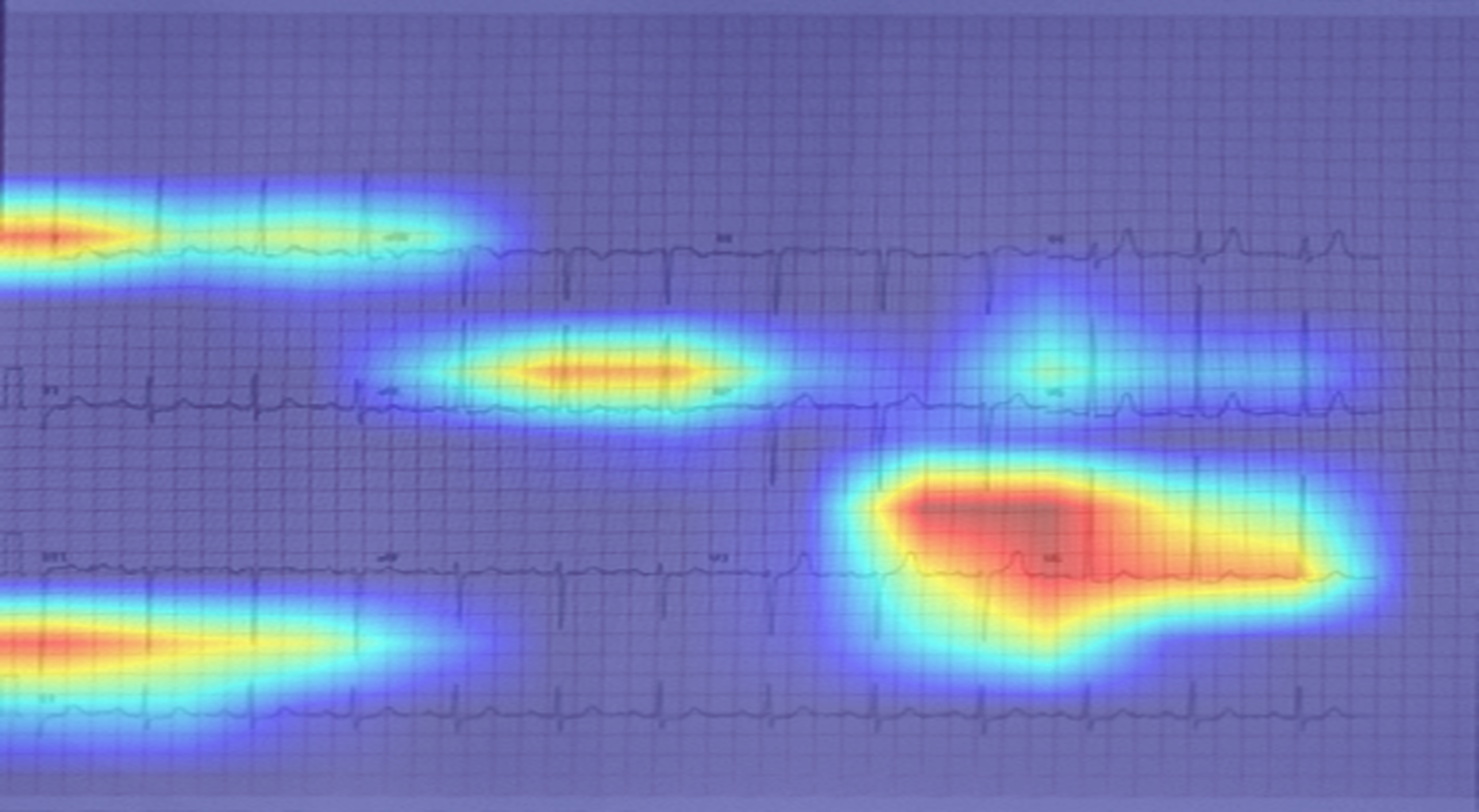}
        \caption{HYP: Full Pipeline (Seg. + Rect.)}
        \label{fig:all_hyp_full}
    \end{subfigure}
    
    \caption{Grad-CAM visualizations for different pre-processing strategies on two samples: Atrial Fibrillation (AF, Top Row, a-c) and Hypertrophy (HYP, Bottom Row, d-f).
    (a, d) The Raw Input setting focuses on background noise.
    (b, e) The ROI Crop + CLAHE model shows a balanced focus that includes rhythm, which is the correct strategy for AF (b).
    (c, f) The Full Pipeline (Seg. + Rect.) model strongly prioritizes single-beat morphology. While this morphological focus is accurate for HYP (f), this bias toward morphology is less effective for the rhythm-based disorder AF (c).}
    \label{fig:gradcam_all}
\end{figure*}

The Grad-CAM visualisations (Fig.~\ref{fig:gradcam_all}) show that different pre-processing methods encourage the model to focus on different scales of information:

\begin{itemize}
\item \textbf{ROI Crop + CLAHE preserves global context:} 
Because this method retains the original layout and avoids aggressive transformations, the model processes the ECG as a continuous temporal signal. 
Figure~\ref{fig:gradcam_all}b shows extended horizontal activation along the rhythm strip, indicating attention to \textbf{temporal continuity}.
This is the appropriate strategy for detecting rhythm disorders such as Atrial Fibrillation (AF).
\item \textbf{Full Pipeline (Seg. + Rect.) emphasises local morphology:} 
By segmenting and rectifying the image, this method enhances individual beats but reduces temporal information. 
As shown in Fig.~\ref{fig:gradcam_all}c, the model concentrates on single-beat morphology. 
This benefits morphology-driven conditions such as Hypertrophy (Fig.~\ref{fig:gradcam_all}f), but weakens rhythm analysis, resulting in lower performance on AF.
\end{itemize}

% --- Figure 1: Grad-CAM Comparison ---

\subsubsection{Robustness vs Complexity: The Trade-off}

\begin{figure*}[t!]
    \centering
    
    \begin{subfigure}[b]{0.23\textwidth}
        \centering
        \includegraphics[width=\textwidth]{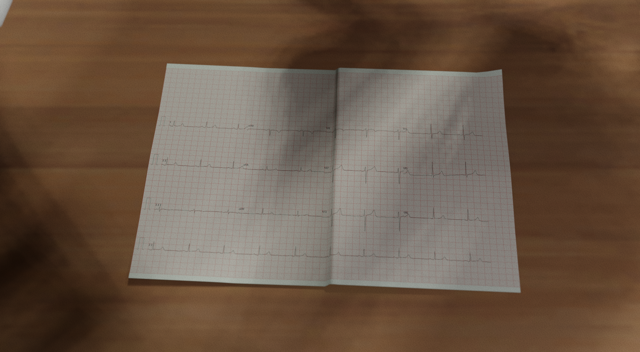}
        \caption{Original}
    \end{subfigure}
    \hfill
    \begin{subfigure}[b]{0.23\textwidth}
        \centering
        \includegraphics[width=\textwidth]{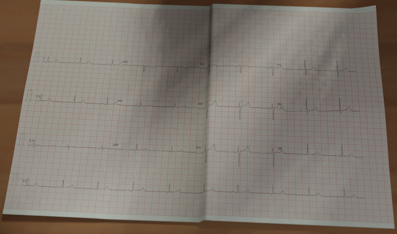}
        \caption{Crop Only}
    \end{subfigure}
    \hfill
    \begin{subfigure}[b]{0.23\textwidth}
        \centering
        \includegraphics[width=\textwidth]{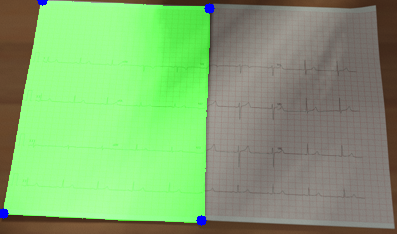}
        \caption{Segmentation Mask}
    \end{subfigure}
    \hfill
    \begin{subfigure}[b]{0.23\textwidth}
        \centering
        \includegraphics[width=\textwidth]{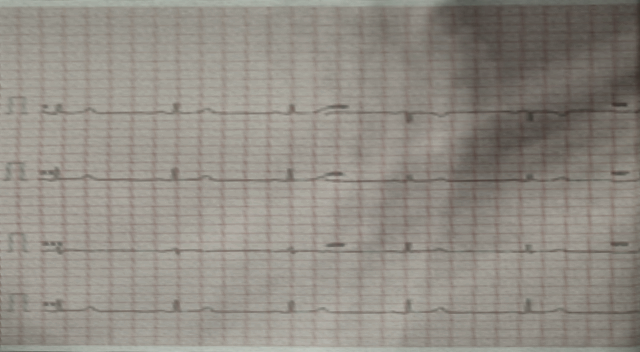}
        \caption{Final Output}
    \end{subfigure}
    
    \vspace{1em}
    
    \begin{subfigure}[b]{0.23\textwidth}
        \centering
        \includegraphics[width=\textwidth]{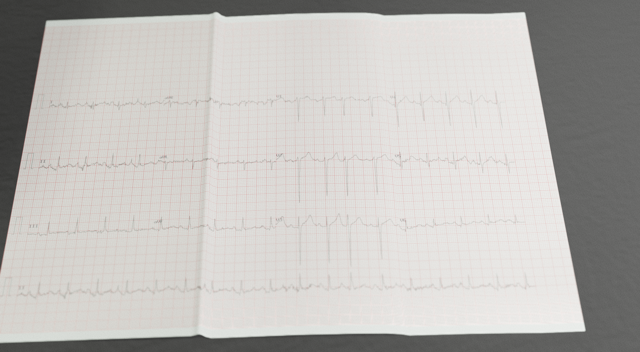}
        \caption{Original}
    \end{subfigure}
    \hfill
    \begin{subfigure}[b]{0.23\textwidth}
        \centering
        \includegraphics[width=\textwidth]{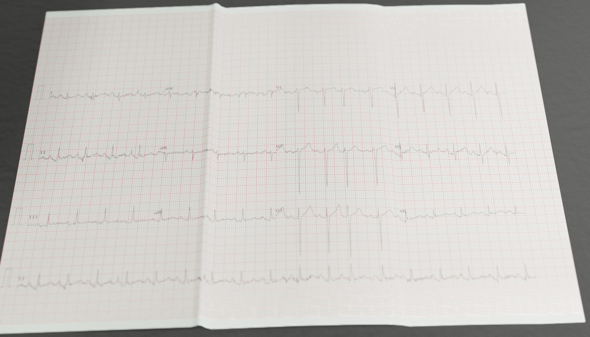}
        \caption{Crop Only}
    \end{subfigure}
    \hfill
    \begin{subfigure}[b]{0.23\textwidth}
        \centering
        \includegraphics[width=\textwidth]{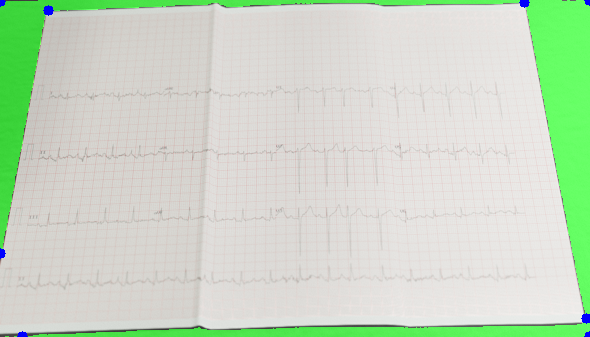}
        \caption{Segmentation Mask}
    \end{subfigure}
    \hfill
    \begin{subfigure}[b]{0.23\textwidth}
        \centering
        \includegraphics[width=\textwidth]{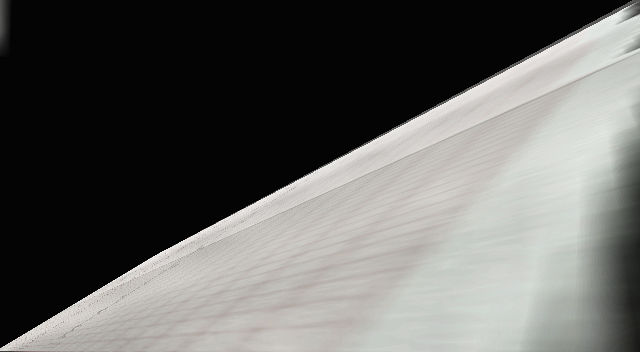}
        \caption{Final Output}
    \end{subfigure}
    
    \caption{Visual analysis of robustness failures in the Full Pipeline (Seg. + Rect.) strategy. 
    \textbf{Top Row (Sample 170):} The segmentation algorithm mistakenly identifies a paper crease as the document boundary (c), resulting in a cropped image that contains only half of the ECG data (d). 
    \textbf{Bottom Row (Sample 435):} The algorithm fails to correctly detect the paper corner (c), causing severe perspective distortion in the final output (d).}
    \label{fig:seg_failure}
\end{figure*}

The Full Pipeline (Seg. + Rect.) represents a "high-risk, high-reward" approach that depends on accurate segmentation and perspective correction. 
However, as illustrated in Fig.~\ref{fig:seg_failure}, this complexity also introduces fragility.

When handling synthetic images with common artefacts (e.g., creases, shadows), the segmentation step may fail catastrophically. 
As shown in Sample 170 (in the top row of Fig.~\ref{fig:seg_failure}), a crease-induced shadow is misinterpreted as a document boundary, causing the algorithm to retain only half of the ECG trace and producing a severely corrupted input. 
Likewise, in Sample 435 (in the bottom row of Fig.~\ref{fig:seg_failure}), the failure to detect one of the document's corners leads to an incorrect perspective transformation, resulting in extreme geometric distortion that renders the final image unusable. 
These examples illustrate how fragile the segmentation process becomes under realistic artefacts, ultimately hindering reliable model training.

In contrast, ROI Crop + CLAHE is a low-intervention strategy that prioritises data integrity over geometric precision. 
Our results indicate that modern CNNs are naturally robust to perspective distortion (e.g., tilted images), but are highly sensitive to missing waveform information caused by segmentation failures. 
Preserving the completeness of the ECG therefore proves more important than achieving perfect geometric alignment.

These robustness challenges imply that the artefacts introduced by the Full Pipeline (Seg. + Rect.) frequently outweigh the potential benefits of geometric normalisation. By contrast, the ROI Crop + CLAHE strategy, which omits these fragile operations, provides a more reliable balance: it consistently removes background interference while preserving the structural integrity of the synthetic ECG data. For this reason, ROI Crop + CLAHE is adopted as the standard pre-processing module within our framework.

% --- Figure 2: Robustness Failure ---
\section{Conclusion}
\label{sec:conclusion}

In this study, we addressed the "Single-Source Limitation" commonly observed when classifying synthetic ECG photographs. 
To overcome this issue, we proposed a Knowledge Augmentation (KA) framework that uses heterogeneous synthetic data (specifically, synthetic scan-like images) to improve performance on photo-based ECG inputs.

Our work offers two main contributions. 
First, we developed a robust pre-processing pipeline that reduces background noise and brings greater consistency to visual inputs. 
Second, we introduced a two-stage training strategy consisting of a Morphology Learning Stage and a Task-Specific Adaptation Stage. 
This design enables the model to learn general morphological patterns from heterogeneous sources before adapting to the characteristics of ECG photographs.
Our experiments show that this approach clearly outperforms traditional single-source training methods. 
The effectiveness of the proposed framework was further confirmed by achieving first place in the 2024 BHF Challenge.

For future work, we plan to enhance the model by exploring techniques that can better capture \textbf{fine-grained image details}. 
Such improvements may allow the model to recognise subtle morphological differences in ECG images, which are essential for reliable clinical interpretation. 
In addition, evaluating this framework on \textbf{real-world, non-synthetic} photographic ECG datasets will be an important step towards assessing its generalisation ability in clinical environments.

\section*{Declaration of generative AI and AI-assisted technologies in the manuscript preparation process}

During the preparation of this work the author(s) used Google Gemini in order to improve the language and readability of the manuscript. After using this tool/service, the author(s) reviewed and edited the content as needed and take(s) full responsibility for the content of the published article.

% \begin{thebibliography}{00}

% %% For authoryear reference style
% %% \bibitem[Author(year)]{label}
% %% Text of bibliographic item

% \bibitem[Lamport(1994)]{lamport94}
%   Leslie Lamport,
%   \textit{\LaTeX: a document preparation system},
%   Addison Wesley, Massachusetts,
%   2nd edition,
%   1994.

% \end{thebibliography}

\bibliographystyle{cas-model2-names}
\bibliography{cas-refs}
\end{document}